\documentclass[final]{cvpr}

\usepackage{times}
\usepackage{epsfig}
\usepackage{graphicx}
\usepackage{amsmath}
\usepackage{amssymb}
\usepackage{bbm}
\usepackage{tabularx} 
\usepackage{multirow}
\usepackage{multicol}
\usepackage{floatrow}

\usepackage[pagebackref=true,breaklinks=true,colorlinks,bookmarks=false]{hyperref}

\def\cvprPaperID{****} 
\def\confYear{CVPR 2021}

\begin{document}

\title{Data-Efficient Language-Supervised Zero-Shot Learning with Self-Distillation}

\author{Ruizhe Cheng$^1$, Bichen Wu$^2$, Peizhao Zhang$^2$, Peter Vajda$^2$, Joseph E. Gonzalez$^1$\\
$^1$UC Berkeley, $^2$Facebook Reality Labs\\
{\tt\small \{chengruizhe,jegonzal\}@berkeley.edu, \{wbc,stzpz,vajdap\}@fb.com}



}

\maketitle

\begin{abstract}
Traditional computer vision models are trained to predict a fixed set of predefined categories. Recently, natural language has been shown to be a broader and richer source of supervision that provides finer descriptions to visual concepts than supervised "gold" labels. Previous works, such as CLIP, use a simple pretraining task of predicting the pairings between images and text captions. CLIP, however, is data hungry and requires more than 400M image text pairs for training. We propose a data-efficient contrastive distillation method that uses soft labels to learn from noisy image-text pairs. Our model transfers knowledge from pretrained image and sentence encoders and achieves strong performance with only 3M image text pairs, 133x smaller than CLIP. Our method exceeds the previous SoTA of general zero-shot learning on ImageNet 21k+1k by 73\% relatively with a ResNet50 image encoder and DeCLUTR text encoder. We also beat CLIP by 10.5\% relatively on zero-shot evaluation on Google Open Images (19,958 classes).
\vspace{-0.5cm}
\end{abstract}

\section{Introduction}
In real-world image recognition tasks, input images can come from a broad range of distributions, spanning tens of thousands of object categories unknown during training. It is thus important for computer vision models to generalize to a large number of visual concepts that may or may not be present in the training data. This problem is called zero-shot learning (ZSL), which aims to transfer knowledge from some known classes with training data to a much larger number of unfamiliar classes. In this paper, we focus on the general zero-shot learning scenario where, at test time, the labels can be either seen or unseen classes. \\

\begin{figure}[t!]
  \centering
  \includegraphics[width=0.8\columnwidth]{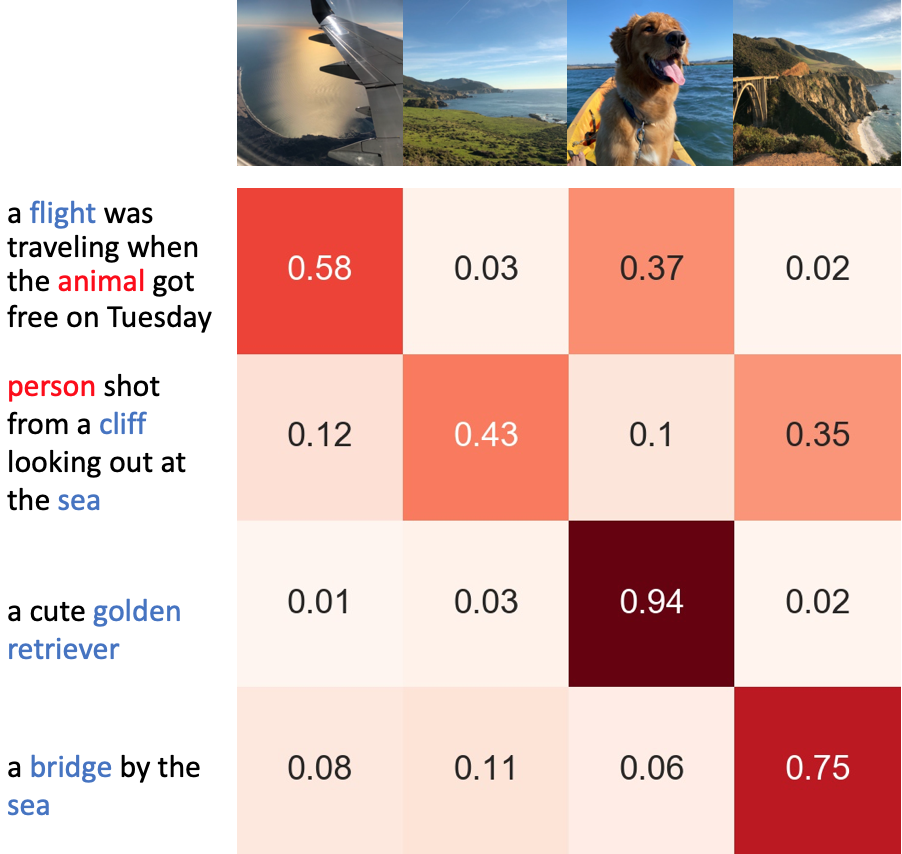}
  \caption{Caption and image pairings are noisy.  Images may contain objects not mentioned in the caption, and captions have words not related to the image (colored red). There is a many-to-many relationship between a batch of images and captions, which is better modeled by soft probabilities than hard labels.  Self-distillation with soft labels mitigates this noise, and enables us to achieve good performance with high data efficiency.}
  \label{fig:demo}
\end{figure}

Traditional ZSL methods mainly follow three paradigms. The first paradigm uses pretrained word embedding vectors to represent different categories and implicitly model their relationships. DeViSE\cite{devise} projects image features from a pretrained CNN and label's word embeddings into a common embedding space. ConSE\cite{conse} proposes a convex combination of the top k most likely image embeddings. The second explicitly models class relationships as a graph, and use a graph convolutional network (GCN), or a predefined class hierarchy, such as WordNet\cite{wordnet}, to learn the knowledge propagation between classes. GCNZ\cite{gcn} and DGPZ\cite{dgp} use a GCN to propagate knowledge into classifiers of unseen classes, while using CNN and word embeddings to encode image and label features. HZSL\cite{hyperbolic} projects image and text embeddings into a hyperbolic space that groups together child and parent classes in the WordNet\cite{wordnet} class hierarchy. Lastly, \cite{embarrassing, fined_grained, label_embedding} rely on human-labeled attributes to model semantics of classes.


These works, however, have several drawbacks. First, they focus on finding a better mapping between image features extracted from pretrined CNNs and pretrained word embeddings such as GloVe\cite{glove}. The image and text embeddings are not trained end-to-end jointly, limiting the generalization power and the quality of feature representations. Second, predefined class hierarchies, such as WordNet\cite{wordnet}, model categories in a tree structure, which fails to capture the complicated inter-class relationships present in real-world objects. Third, reliance on class hierarchies also limits the scope of classifiable objects to those present in the hierarchy. Fourth, methods that depend on attributes cannot generalize to categories that do not have known attributes. 

In recent years, natural language has become a powerful source of supervision for image representation learning. \cite{instagram} shows that pretraining by predicting hashtags on Instagram improves performance on ImageNet by over 5\%. \cite{virtex, icmlm, convirt} all demonstrate the effectiveness of transformer-based language modeling in learning image representation from text. CLIP\cite{clip} and ALIGN\cite{align} apply natural language supervision to the domain of ZSL. CLIP collects an enormous dataset with over 400M image caption pairs from the Internet, and trains an image encoder and a text encoder jointly with a contrastive loss to maximize the cosine similarity of corresponding image text embeddings and minimize those of others. CLIP demonstrates good zero-shot classification results on 27 downstream image classification datasets. However, neither CLIP nor ALIGN has published their image-caption datasets. It's also an expensive and daunting task to collect, maintain and train vision models on datasets of that size. \\

We propose a data-efficient ZSL training pipeline that enables any pretrained image encoders to generalize to unseen classes. We initialize our model with an image encoder pretrained on ImageNet\cite{imagenet} 1k and a pretrained universal sentence encoder. We train our models on the public Conceptual Captions\cite{cc} dataset, which contains 3M loosely correlated image caption pairs. As seen in figure \ref{fig:demo}, there is considerable noise in the image-text pairings collected from the Internet. CLIP uses hard labels in the contrastive loss and account for the noise with a lot of data. Instead, we propose to use a hybrid of hard contrastive and soft distillation losses. We distill the model from its running Exponential Moving Average(EMA) with soft labels, as a method of denoising. Learning from soft labels enables better modelling of the rich correlations between vision and language and effectively account for cases where one caption matches objects in multiple images and vice versa. EMA is used as a continuous version of repeated self-distillation \cite{distillation, label_refinery}.

With a ResNet50\cite{resnet} image encoder and DeCLUTR\cite{declutr} text encoder, we outperform the current SoTA of general ZSL on ImageNet 21k+1k by 73\% relatively. In addition, we recognize issues with ImageNet21k and the 27 datasets used by CLIP\cite{clip} for ZSL evaluation in section \ref{sec:eval_dataset}. To bypass these problems, we propose using Google Open Images\cite{goi}, which contains 19,958 categories, as a benchmark for zero-shot knowledge transfer to common visual concepts. Our model also exceeds CLIP on GOI by 10.5\% relatively, while using a $>$100x smaller dataset.\\
\vspace{-0.5cm}

\begin{table*} \label{tab:GOI}
\begin{center}
\begin{tabularx}{0.899\textwidth}{c|c|c|c|c|c|c|c|c|c}
\hline
\multirow{2}{*}{Dataset} & \multirow{2}{*}{Size} & \multirow{2}{*}{Model} & \multirow{2}{*}{Image Encoder} & \multirow{2}{*}{Text Encoder} & \multirow{2}{*}{Params} & \multicolumn{4}{c}{Flat Hit@k(\%)} \\
\cline{7-10}
& & & & & & 1 & 2 & 5 & 10 \\
\hline\hline
CLIP & 400M & CLIP & ResNet50 & Bert Base* & 102M & 26.5 & 38.3 & 54.0 & 64.3\\
CLIP & 400M & CLIP & ViT-B/32 & Bert Base* & 151M & 27.5 & 39.5 & 55.3 & 65.4\\
\hline
CC & 3M & C & FBNet C & DeCLUTR Sci Base & 114M & 20.8 & 31.5 & 47.7 & 60.0 \\
CC & 3M & C & EfficientNet B0 & DeCLUTR Sci Base & 114M & 23.3 & 34.8 & 51.4 & 63.5\\
CC & 3M & C & ResNet50 & Sentence Bert Base & 134M & 22.5 & 33.1 & 47.8 & 58.2\\
CC & 3M & C & ResNet50 & Bert Base & 134M & 24.6 & 35.4 & 50.0 & 60.2\\
CC & 3M & C & ResNet50 & DeCLUTR Sci Base & 135M & 28.2 & 40.6 & 57.6 & 68.7\\
CC & 3M & C+D & ResNet50 & DeCLUTR Sci Base & 135M & \textbf{29.3} & \textbf{42.0} & \textbf{58.6} & \textbf{69.4}\\
\hline
\end{tabularx}
\end{center}
\caption{Flat hit @k on Google Open Images. In the Model column, C means trained using contrastive loss only, and C+D means trained with contrastive and distillation loss jointly. * means that the model is a modified version.}
\vspace{-0.5cm}
\end{table*}

\section{Methods}
Our model has a two-tower structure with an image encoder and a sentence encoder that outputs fixed-sized embeddings for a batch of corresponding images and captions. Different from pervious ZSL works, our model assumes no class hierarchy. This makes our method more general, and easily extensible to datasets like Google Open Images\cite{goi}.

\subsection{Visual and Language Pretraining}
Pretraining has become a crucial procedure in many NLP tasks\cite{bert, gpt3, roberta}. Likewise, BiT\cite{BiT} and ViT\cite{ViT} has shown that transfer of pretrained visual representations leads to significant performance gains. Therefore, we initialize our model with an image encoder pretrained on ImageNet\cite{imagenet} 1k and a pretrained universal sentence encoder, such as Sentence Transformers\cite{sentence_transformer} or DeCLUTR\cite{declutr}. Sentence Transformers are pretrained on SNLI\cite{SNLI} and MultiNLI\cite{MultiNLI}, whereas DeCLUTR is pretrained on the OpenWebText Corpus\cite{OpenWebText} or the Semantic Scholar Open Research Corpus\cite{s2orc}. 

\subsection{Contrastive Learning}
The contrastive learning\cite{contrastive} objective has been widely used in NLP and is at the core of several unsupervised\cite{efficient_cpc, unsupervised_feature_learning, mutual_info_maximization} and self-supervised learning works\cite{moco, simclr}. Similar to CLIP\cite{clip}, we also use the contrastive loss, which measures the similarities of sample pairs in an embedding space. Specifically, we use the InfoNCE\cite{cpc} loss where similarity is measured by dot product. Take a batch of $N$ image and text pairs, the image and text encoders are joinly trained to maximize the cosine similarity of the $N$ positive image and text pairings while minimizing the cosine similarity of the other $N^2 - N$ negative image text pairings. In a batch of $N$ image text pairs, let $z_i^I$ be the embedding of the $i$th image, and $z_j^T$ that of the $j$th text. The probability of the $i$th image matching the $j$th text is: 

\begin{equation}\label{eq:1}
P(z_i^I, z_j^T; \tau) = \frac{\exp(z_i^I \cdot z_{j}^T / \tau)}{\sum_{k=0}^{N}\exp(z_i^I \cdot z_k^T / \tau)}
\end{equation}

The InfoNCE loss for images is defined as:

\begin{equation} \label{eq:2}
L_I = - \frac{1}{N}\sum_{i=0}^{N}\log P(z_i^I, z_i^T; \tau)
\end{equation}

We define the probability in (\ref{eq:1}) similarly for texts, and compute the InfoNCE loss symmetrically to get $L_T$. The contrastive loss function thus becomes:
\begin{equation} \label{eq:3}
L_{\mbox{InfoNCE}} = \frac{1}{2}(L_I + L_T)
\end{equation}

\subsection{EMA Self-Distillation}
Image-text pairs collected from the Internet are usually only weakly correlated and noise is abundant. Often, images contain objects not mentioned in their captions, and captions contain words unrelated to their images. It's also common for one caption to match objects in multiple images in a single batch. Hence, it's not ideal to use hard labels as the only learning objective. We keep an Exponential Moving Average(EMA) of our model during training and use it as a continuously evolving teacher for self-distillation. We use a KL divergence loss to match the outputs of our model and its EMA teacher. According to equation (\ref{eq:1}), define $P_M^I$ and $P_{EMA}^I$ as the probability distribution of images over texts in a batch, for our model and its EMA teacher, respectively. Symmetrically, define $P_M^T$ and $P_{EMA}^T$.
\vspace{-0.1cm}
\begin{equation}
L_{\mbox{KL}} = \frac{1}{2}[\mbox{KL}(P_M^I, P_{EMA}^I) + \mbox{KL}(P_M^T, P_{EMA}^T)]
\end{equation}
The final loss we use is:
\vspace{-0.1cm}
\begin{equation}
L = L_{\mbox{InfoNCE}} + \alpha L_{\mbox{KL}}
\end{equation}
where $\alpha$ is set to 1.0 in our experiments.

\begin{table*} \label{tab:IN22k}
\begin{center}
\begin{tabularx}{0.902\textwidth}{c|c|c|c|c|c|c|c|c}
\hline
\multirow{2}{*}{Dataset} & \multirow{2}{*}{Size} & \multirow{2}{*}{Model} & \multirow{2}{*}{Image Encoder} & \multirow{2}{*}{Text Encoder} & \multicolumn{4}{c}{Flat Hit@k(\%)} \\
\cline{6-9}
& & & & & 1 & 2 & 5 & 10 \\
\hline\hline
ImageNet1k & 1.2M & DeViSE & ResNet50 & skip-gram & 0.3 & 0.9 & 2.2 & 3.6\\
ImageNet1k & 1.2M & ConSE & ResNet50 & skip-gram & 0.1 & 1.5 & 3.5 & 4.9\\
ImageNet1k & 1.2M & GCNZ & ResNet50 & GloVe & 1.0 & 2.3 & 5.3 & 8.1\\
ImageNet1k & 1.2M & HZSL & ResNet50 & GloVe* & 2.2 & 4.6 & 9.2 & 12.7\\
\hline
CC & 3M & C & FBNet C\cite{fbnet} & DeCLUTR Sci Base & 2.7 & 4.0 & 7.5 & 11.1\\
CC & 3M & C & EfficientNet B0\cite{efficientnet} & DeCLUTR Sci Base & 3.0 & 4.6 & 8.4 & 12.2\\
CC & 3M & C & ResNet50 & Bert Base & 3.2 & 5.7 & 10.5 & 15.3\\
CC & 3M & C & ResNet50 & Sentence Bert Base & 3.6 & 5.4 & 10.1 & 14.7\\
CC & 3M & C & ResNet50 & DeCLUTR Sci Base & 3.8 & 5.5 & 9.8 & 13.9\\
CC & 3M & C+D & ResNet50 & DeCLUTR Sci Base & 3.7 & 5.4 & 9.5 & 13.6\\
CC & 3M & C & ViT-Deit-B/16\cite{vit_deit} & DeCLUTR Sci Base & \textbf{4.0} & \textbf{6.0} & \textbf{10.9} & \textbf{15.5}\\
\hline
CLIP & 400M & CLIP & ResNet50 & Bert Base* & 13.5 & 19.7 & 30.5 & 39.4\\
CLIP & 400M & CLIP & ViT-B/32 & Bert Base* & 15.3 & 22.2 & 33.9 & 43.3\\
\hline
\end{tabularx}
\caption{Flat hit @k on ImageNet 21k+1k.}
\end{center}
\vspace{-0.5cm}
\end{table*}

\section{Experiments}

\subsection{Training}
We apply a training schedule similar to the finetuning step of BiT\cite{BiT}. We use SGD with an initial learning rate of 3e-3, a cosine annealing lr scheduler, momentum 0.9, and no weight decay. Input images are resized to 256x256 and random cropped to 224x224. We train the model on 4 GPUs using Pytorch\cite{pytorch} Distributed Data Parallel with a batch size of 128 per GPU for 30 epochs. While CLIP\cite{clip} computes the contrastive loss using only the batch on each GPU, we find that it's important to all gather logits from the other GPUs and use them as negative samples.

\subsection{Evaluation}
During evaluation,we use a prompt template of ``a photo of \{label\}" to augment the text labels of the target categories. We then compute the text embeddings of test categories with the trained text encoder, and fit a KNN using the embeddings. Given an image, we find the top k nearest neighors of its embedding based on cosine similarity.

\subsubsection{Evaluation Metric}
The main metric we use for evaluating performance of ZSL is flat hit@k. Flat hit@k is the percentage of test images such that the top k predictions the model returns overlaps with any of the true labels. In ImageNet\cite{imagenet}, each image is only labeled with one synset, but in Google Open Images\cite{goi}, each image is labeled with multiple classes. The formal definition of flat hit@k is:
\vspace{-0.2cm}
\begin{equation} \label{eq:4}
\mbox{flat hit@k} = \frac{1}{N} \sum_{i=1}^{N} \mathbbm{1}\{ \{F(x_i)\}_K \cap L_i \neq \varnothing \}
\vspace{-0.2cm} 
\end{equation}
where $\{F(x_i)\}_K $ is the top k predictions for the $i$th image and $L_i$ is the set of true labels.

\subsubsection{Evaluation Dataset}
 \label{sec:eval_dataset}
We measure the ZSL performance mainly on Google Open Images \cite{goi}. And for backward compatibility to compare with prior work, we also report the results on ImageNet 21K+1K benchmark. We do not report results on the 27 datasets benchmark used by CLIP\cite{clip}. We discuss our considerations below.

\textbf{ImageNet 21K+1K:} Despite its popularity, there are four main problems of using ImageNet\cite{imagenet} for ZSL evaluation. First, based on the WordNet\cite{wordnet} structure, ImageNet has many repeated or trivially different classes. For example, "sunglass" and "sunglasses" are two different classes. Out of 22843 synsets, 1128 of them have names identical to at least another synset. Second, ImageNet labels don't distinguish words with multiple meanings. For example, the word "crane" can mean either a type of bird or machine. Both classes are in ImageNet but have the same label. This happens for many words such as "ball". Third, each image in ImageNet is only labeled with exactly one class. When there are 2 or more visual concepts in the image, the model is forced to guess which object to classify. 
Fourth, ImageNet lacks the interactions between different visual concepts. About 90\% of the images in ImageNet have only 1 distinct class, and almost no images have more than 4 distinct classes. \\

\vspace{-0.4cm}
\textbf{Google Open Image:} Compared to ImageNet, Google Open Images\cite{goi} also contains a wide range of concepts, and it fixes all four problems outlined above. There are no repeated labels for different classes in GOI. Words with multiple meanings are also differentiated. For example, "crane" is labeled with ``Crane (Machine)" and ``Crane (Bird)". More importantly, GOI labels each image with multiple classes, largely eliminating false negatives. In addition, GOI contains much more interactions between distinct classes per image, where more than 60\% of images have 2 or more distinct classes. Inter-class interactions are especially useful in zero-shot learning, when we aim to transfer knowledge from seen to unseen classes.

\textbf{CLIP benchmark with 27 datasets:}
CLIP\cite{clip} evaluates their model on 27 image classification datasets. However, many of these datasets are domain specific, such as Stanford Cars and FGVC aircraft, which have specific models of cars or planes as categories. This makes evaluation on them a test a knowledge memorization, rather than generalization. Similar to ImageNet, very few of these datasets contain multiple distinct classes in the same image, reflecting a lack of visual richness. Lastly, with only 3896 total categories, the 27 datasets altogether don't cover nearly as many common visual concepts as GOI.

\subsection{Results on Google Open Images}
We evaluate the models on the test set of Google Open Images V6\cite{goi}, with 125,436 images. Traditional ZSL baselines aren't evaluated on GOI due to the lack of a class structure. In table \ref{tab:GOI}, we compare the flat hit@k of our models with pretrained CLIP\cite{clip}. Our ResNet50 and DeCLUTR Sci Base model trained with the joint contrastive and distillation loss exceeds CLIP ResNet50 and Bert\cite{bert} by 10.5\% relatively in FH@k=1, while being $>$ 100x more data efficient.

\subsection{Results on ImageNet 21k+1k}
In this section, we present flat hit@k results on zero-shot transfer to the ImageNet 21k+1k\cite{imagenet} dataset, which contains 21841 classes in total. The image encoders are initialized with weights pretrained on ImageNet 1k. Sentence Bert\cite{sentence_transformer} is pretrained on SNLI\cite{SNLI} and MultiNLI\cite{MultiNLI}, while Declutr Sci Base\cite{declutr} is pretrained on the S2ORC\cite{s2orc}. \\

Many traditional ZSL methods rely on a predefined class hierarchy for explicit knowledge propagation. ImageNet, whose classes are a subset of WordNet, becomes the ideal benchmark for these works. With 400M image text pairs, CLIP\cite{clip} vastly outperforms previous methods. Our method uses Conceptual Captions\cite{cc} 3M, which is on the same order of magnitude as ImageNet 1k, and outperforms the previous SoTA, HZSL\cite{hyperbolic}, by 73\% relatively. In table \ref{tab:IN22k}, we demonstrate good performance on a variety of image and sentence encoder architectures. The gap between our method and CLIP may be caused by the fact that ImageNet classes contain many uncommon words, such as scientific names of animals or medical terms. CLIP's dataset is much larger and thus covers much more uncommon words. EMA distillation also slightly decreases the performance compared to using only contrastive loss. We hypothesize that this is because ImageNet only has one "gold" label per image during evaluation. However, EMA distillation encourages the model to output a softer probability output for multiple classes, which can be present but just not labeled.

{\small
\bibliographystyle{ieee_fullname}
\bibliography{egbib}

\begin{thebibliography}{10}\itemsep=-1pt

\bibitem{label_embedding}
Zeynep Akata, Florent Perronnin, Zaid Harchaoui, and Cordelia Schmid.
\newblock Label-embedding for attribute-based classification.
\newblock {\em CVPR}, 2013.

\bibitem{fined_grained}
Zeynep Akata, Scott Reed, Daniel Walter, Honglak Lee, and Bernt Schiele.
\newblock Evaluation of output embeddings for fine-grained image
  classification.
\newblock {\em CVPR}, 2015.

\bibitem{label_refinery}
Hessam Bagherinezhad, Maxwell Horton, Mohammad Rastegari, and Ali Farhadi.
\newblock Label refinery: Improving ima- genet classification through label
  progression.
\newblock {\em arXiv preprint arXiv:1805.02641}, 2018.

\bibitem{SNLI}
Samuel~R. Bowman, Gabor Angeli, Christopher Potts, and Christopher~D. Manning.
\newblock A large annotated corpus for learning natural language inference.
\newblock 2015.

\bibitem{gpt3}
Tom~B. Brown, Benjamin Mann, Nick Ryder, Melanie Subbiah, Jared Kaplan,
  Prafulla Dhariwal, Arvind Neelakantan, Pranav Shyam, Girish Sastry, Amanda
  Askell, Sandhini Agarwal, Ariel Herbert-Voss, Gretchen Krueger, Tom Henighan,
  Rewon Child, Aditya Ramesh, Daniel~M. Ziegler, Jeffrey Wu, Clemens Winter,
  Christopher Hesse, Mark Chen, Eric Sigler, Mateusz Litwin, Scott Gray,
  Benjamin Chess, Jack Clark, Christopher Berner, Sam McCandlish, Alec Radford,
  Ilya Sutskever, and Dario Amodei.
\newblock Language models are few-shot learners.
\newblock {\em arXiv preprint arXiv:2005.14165}, 2020.

\bibitem{simclr}
Ting Chen, Simon Kornblith, Mohammad Norouzi, and Geoffrey Hinton.
\newblock A simple framework for contrastive learning of visual
  representations.
\newblock {\em ICML}, 2020.

\bibitem{imagenet}
Jia Deng, Wei Dong, Richard Socher, Li-Jia Li, Kai Li, and Li Fei-Fei.
\newblock Imagenet: A large-scale hierarchical image database.
\newblock pages 248--255, 2009.

\bibitem{virtex}
Karan Desai and Justin Johnson.
\newblock Virtex: Learning visual representations from textual annotations.
\newblock {\em arXiv preprint arXiv:2006.06666}, 2020.

\bibitem{bert}
Jacob Devlin, Ming-Wei Chang, Kenton Lee, and Kristina Toutanova.
\newblock Bert: Pre-training of deep bidirectional transformers for language
  understanding.
\newblock {\em arXiv preprint arXiv:1810.04805}, 2018.

\bibitem{ViT}
Alexey Dosovitskiy, Lucas Beyer, Alexander Kolesnikov, Dirk Weissenborn,
  Xiaohua Zhai, Thomas Unterthiner, Mostafa Dehghani, Matthias Minderer, Georg
  Heigold, Sylvain Gelly, Jakob Uszkoreit, and Neil Houlsby.
\newblock An image is worth 16x16 words: Transformers for image recognition at
  scale.
\newblock {\em arXiv preprint arXiv:2010.11929}, 2020.

\bibitem{wordnet}
Ingo Feinerer and Kurt Hornik.
\newblock {\em wordnet: WordNet Interface}, 2020.
\newblock R package version 0.1-15.

\bibitem{devise}
Andrea Frome, Greg~S. Corrado, Jonathon Shlens, Samy Bengio~Jeffrey Dean,
  Marc’Aurelio Ranzato, and Tomas Mikolov.
\newblock Devise: A deep visual-semantic embedding model.
\newblock {\em NIPS}, 2013.

\bibitem{declutr}
John~M Giorgi, Osvald Nitski, Gary~D. Bader, and Bo Wang.
\newblock Declutr: Deep contrastive learning for unsupervised textual
  representations.
\newblock {\em arXiv preprint arXiv:2006.03659}, 2020.

\bibitem{OpenWebText}
Aaron Gokaslan, Vanya Cohen, Ellie Pavlick, and Stefanie Tellex.
\newblock Openwebtext corpus, 2019.

\bibitem{contrastive}
Raia Hadsell, Sumit Chopra, and Yann LeCun.
\newblock Dimensionality reduction by learning an invariant mapping.
\newblock {\em CVPR}, 2006.

\bibitem{moco}
Kaiming He, Haoqi Fan, Yuxin Wu, Saining Xie, and Ross Girshick.
\newblock Momentum contrast for unsupervised visual representation learning.
\newblock {\em CVPR}, 2020.

\bibitem{resnet}
Kaiming He, Xiangyu Zhang, Shaoqing Ren, and Jian Sun.
\newblock Deep residual learning for image recognition.
\newblock {\em CVPR}, 2016.

\bibitem{distillation}
Geoffrey Hinton, Oriol Vinyals, and Jeffrey Dean.
\newblock Distilling the knowledge in a neural network.
\newblock In {\em NIPS Deep Learning and Representation Learning Workshop},
  2015.

\bibitem{mutual_info_maximization}
R~Devon Hjelm, Alex Fedorov, Samuel Lavoie-Marchildon, Karan Grewal, Adam
  Trischler, and Yoshua Bengio.
\newblock Learning deep representations by mutual information estimation and
  maximization.
\newblock {\em ICLR}, 2019.

\bibitem{efficient_cpc}
Olivier~J. Hénaff, Aravind Srinivas, Jeffrey~De Fauw, Ali Razavi, Carl
  Doersch, S.~M.~Ali Eslami, and Aaron van~den Oord.
\newblock Data-efficient image recognition with contrastive predictive coding.
\newblock {\em ICML}, 2020.

\bibitem{align}
Chao Jia, Yinfei Yang, Ye Xia, Yi-Ting Chen, Zarana Parekh, Hieu Pham, Quoc~V.
  Le, Yunhsuan Sung, Zhen Li, and Tom Duerig.
\newblock Scaling up visual and vision-language representation learning with
  noisy text supervision.
\newblock {\em arXiv preprint arXiv:2102.05918}, 2020.

\bibitem{dgp}
Michael Kampffmeyer, Yinbo Chen, Xiaodan Liang, Hao Wang, Yujia Zhang, and
  Eric~P. Xing.
\newblock Rethinking knowledge graph propagation for zero-shot learning.
\newblock {\em CVPR}, 2019.

\bibitem{BiT}
Alexander Kolesnikov, Lucas Beyer, Xiaohua Zhai, Joan Puigcerver, Jessica Yung,
  Sylvain Gelly, and Neil Houlsby.
\newblock Big transfer (bit): General visual representation learning.
\newblock {\em arXiv preprint arXiv:1912.11370}, 2020.

\bibitem{goi}
Alina Kuznetsova, Hassan Rom, Neil Alldrin, Jasper Uijlings, Ivan Krasin, Jordi
  Pont-Tuset, Shahab Kamali, Stefan Popov, Matteo Malloci, Alexander
  Kolesnikov, Tom Duerig, and Vittorio Ferrari.
\newblock The open images dataset v4: Unified image classification, object
  detection, and visual relationship detection at scale.
\newblock {\em IJCV}, 2020.

\bibitem{hyperbolic}
Shaoteng Liu, Jingjing Chen, Liangming Pan, Chong-Wah Ngo, Tat-Seng Chua, and
  Yu-Gang Jiang.
\newblock Hyperbolic visual embedding learning for zero-shot recognition.
\newblock {\em CVPR}, 2020.

\bibitem{roberta}
Yinhan Liu, Myle Ott, Naman Goyal, Jingfei Du, Mandar Joshi, Danqi Chen, Omer
  Levy, Mike Lewis, Luke Zettlemoyer, and Veselin Stoyanov.
\newblock Roberta: A robustly optimized bert pretraining approach.
\newblock {\em arXiv preprint arXiv:1907.11692}, 2019.

\bibitem{s2orc}
Kyle Lo, Lucy~Lu Wang, Mark Neumann, Rodney Kinney, and Daniel Weld.
\newblock {S}2{ORC}: The semantic scholar open research corpus.
\newblock In {\em Proceedings of the 58th Annual Meeting of the Association for
  Computational Linguistics}, pages 4969--4983, Online, July 2020. Association
  for Computational Linguistics.

\bibitem{instagram}
Dhruv Mahajan, Ross Girshick, Vignesh Ramanathan, Kaiming He, Manohar Paluri,
  Yixuan Li, Ashwin Bharambe, and Laurens van~der Maaten.
\newblock Exploring the limits of weakly supervised pretraining.
\newblock {\em ECCV}, 2018.

\bibitem{conse}
Mohammad Norouzi, Tomas Mikolov, Samy Bengio, Yoram Singer, Jonathon Shlens,
  Andrea Frome, Greg~S. Corrado, and Jeffrey Dean.
\newblock Zero-shot learning by convex combination of semantic embeddings.
\newblock {\em ICLR}, 2014.

\bibitem{pytorch}
Adam Paszke, Sam Gross, Francisco Massa, Adam Lerer, James Bradbury, Gregory
  Chanan, Trevor Killeen, Zeming Lin, Natalia Gimelshein, Luca Antiga, Alban
  Desmaison, Andreas Kopf, Edward Yang, Zachary DeVito, Martin Raison, Alykhan
  Tejani, Sasank Chilamkurthy, Benoit Steiner, Lu Fang, Junjie Bai, and Soumith
  Chintala.
\newblock Pytorch: An imperative style, high-performance deep learning library.
\newblock In H. Wallach, H. Larochelle, A. Beygelzimer, F. d\textquotesingle
  Alch\'{e}-Buc, E. Fox, and R. Garnett, editors, {\em Advances in Neural
  Information Processing Systems 32}, pages 8024--8035. Curran Associates,
  Inc., 2019.

\bibitem{glove}
Jeffrey Pennington, Richard Socher, and Christopher~D. Manning.
\newblock Glove: Global vectors for word representation.
\newblock {\em EMNLP}, 2014.

\bibitem{clip}
Alec Radford, Jong~Wook Kim, Chris Hallacy, Aditya Ramesh, Gabriel Goh,
  Sandhini Agarwal, Girish Sastry, Amanda Askell, Pamela Mishkin, Jack Clark,
  Gretchen Krueger, and Ilya Sutskever.
\newblock Learning transferable visual models from natural language
  supervision.
\newblock {\em arXiv preprint arXiv:2103.00020}, 2021.

\bibitem{sentence_transformer}
Nils Reimers and Iryna Gurevych.
\newblock Sentence-bert: Sentence embeddings using siamese bert-networks.
\newblock In {\em Proceedings of the 2019 Conference on Empirical Methods in
  Natural Language Processing}. Association for Computational Linguistics, 11
  2019.

\bibitem{embarrassing}
Bernardino Romera-Paredes and Philip H.~S. Torr.
\newblock An embarrassingly simple approach to zero-shot learning.
\newblock {\em ICML}, 2015.

\bibitem{icmlm}
Mert~Bulent Sariyildiz, Julien Perez, and Diane Larlus.
\newblock Learning visual representations with caption annotations.
\newblock {\em arXiv preprint arXiv:2008.01392}, 2020.

\bibitem{cc}
Piyush Sharma, Nan Ding, Sebastian Goodman, and Radu Soricut.
\newblock Conceptual captions: A cleaned, hypernymed, image alt-text dataset
  for automatic image captioning.
\newblock In {\em Proceedings of ACL}, 2018.

\bibitem{efficientnet}
Mingxing Tan and Quoc Le.
\newblock {E}fficient{N}et: Rethinking model scaling for convolutional neural
  networks.
\newblock {\em ICML}, 2019.

\bibitem{vit_deit}
Hugo Touvron, Matthieu Cord, Matthijs Douze, Francisco Massa, Alexandre
  Sablayrolles, and Herv\'e J\'egou.
\newblock Training data-efficient image transformers \& distillation through
  attention.
\newblock {\em arXiv preprint arXiv:2012.12877}, 2020.

\bibitem{cpc}
Aaron van~den Oord, Yazhe Li, and Oriol Vinyals.
\newblock Representation learning with contrastive predictive coding.
\newblock {\em arXiv preprint arXiv:1191.05722}, 2018.

\bibitem{gcn}
Xiaolong Wang, Yufei Ye, and Abhinav Gupta.
\newblock Zero-shot recognition via semantic embeddings and knowledge graphs.
\newblock {\em CVPR}, 2018.

\bibitem{MultiNLI}
Adina Williams, Nikita Nangia, and Samuel Bowman.
\newblock A broad-coverage challenge corpus for sentence understanding through
  inference.
\newblock In {\em Proceedings of the 2018 Conference of the North American
  Chapter of the Association for Computational Linguistics: Human Language
  Technologies, Volume 1 (Long Papers)}, pages 1112--1122. Association for
  Computational Linguistics, 2018.

\bibitem{fbnet}
Bichen Wu, Xiaoliang Dai, Peizhao Zhang, Yanghan Wang, Fei Sun, Yiming Wu,
  Yuandong Tian, Peter Vajda, Yangqing Jia, and Kurt Keutzer.
\newblock Fbnet: Hardware-aware efficient convnet design via differentiable
  neural architecture search.
\newblock {\em CVPR}, 2019.

\bibitem{unsupervised_feature_learning}
Zhirong Wu, Yuanjun Xiong, Stella~X. Yu, and Dahua Lin.
\newblock Unsupervised feature learning via non-parametric instance
  discrimination.
\newblock {\em CVPR}, 2018.

\bibitem{convirt}
Yuhao Zhang, Hang Jiang, Yasuhide Miura, Christopher~D. Manning, and Curtis~P.
  Langlotz.
\newblock Contrastive learning of medical visual representations from paired
  images and texts.
\newblock {\em arXiv preprint arXiv:2010.00747}, 2020.

\end{thebibliography}
}

\end{document}